\documentclass[10pt,twocolumn,letterpaper]{article}

\usepackage[pagenumbers]{cvpr} 
\usepackage[dvipsnames]{xcolor}
\usepackage{colortbl}
\usepackage{xcolor}

\definecolor{cvprblue}{rgb}{0.21,0.49,0.74}
\usepackage[pagebackref,breaklinks,colorlinks,citecolor=cvprblue]{hyperref}

\def\paperID{*****} 
\def\confName{CVPR}
\def\confYear{2024}

\title{NePF: Neural Photon Field for Single-Stage Inverse Rendering}

\author{\textbf{Tuen-Yue Tsui \hspace{0.75cm} Qin Zou}\\
School of Computer Science\\
Wuhan University\\
{\tt\small tsui\_tuenyue@whu.edu.cn} \hspace{0.45cm}
{\tt\small qzou@whu.edu.cn}
}

\begin{document}
\maketitle
\begin{abstract}
We present a novel single-stage framework, Neural Photon Field (NePF), to address the ill-posed inverse rendering from multi-view images. Contrary to previous methods that recover the geometry, material, and illumination in multiple stages and extract the properties from various multi-layer perceptrons across different neural fields, we question such complexities and introduce our method - a single-stage framework that uniformly recovers all properties. NePF achieves this unification by fully utilizing the physical implication behind the weight function of neural implicit surfaces and the view-dependent radiance. Moreover, we introduce an innovative coordinate-based illumination model for rapid volume physically-based rendering. To regularize this illumination, we implement the subsurface scattering model for diffuse estimation. We evaluate our method on both real and synthetic datasets. The results demonstrate the superiority of our approach in recovering high-fidelity geometry and visual-plausible material attributes.
\end{abstract}
\section{Introduction}
\label{sec:intro}

Recovering geometry, material, and illumination properties of an object from multi-view images—referred to as inverse rendering~\cite{10.5555/512416.512422, 10.5555/927098,10.1145/383259.383271}—is a fundamental and persistent problem in both computer vision and computer graphics. The challenge arises from the immense solution set inherent to this problem. Early methods try to leverage additional information~\cite{nerv2021, bi2020nrf, nimierdavid2021material} as prior. In order to tackle these ill-posed problems without prior knowledge, recent techniques adopt a multi-stage strategy or leveraged pre-trained networks~\cite{nerfactor,zhang2023neilf++, boss2021nerd, boss2021neuralpil}, recovering distinct properties at different stages. 
While this multi-stage strategy enables the derivation of plausible results, little attention has been paid to its inherent complexities and limitations.
\begin{figure}[t]
  \centering
   \includegraphics[width=1.0\linewidth]{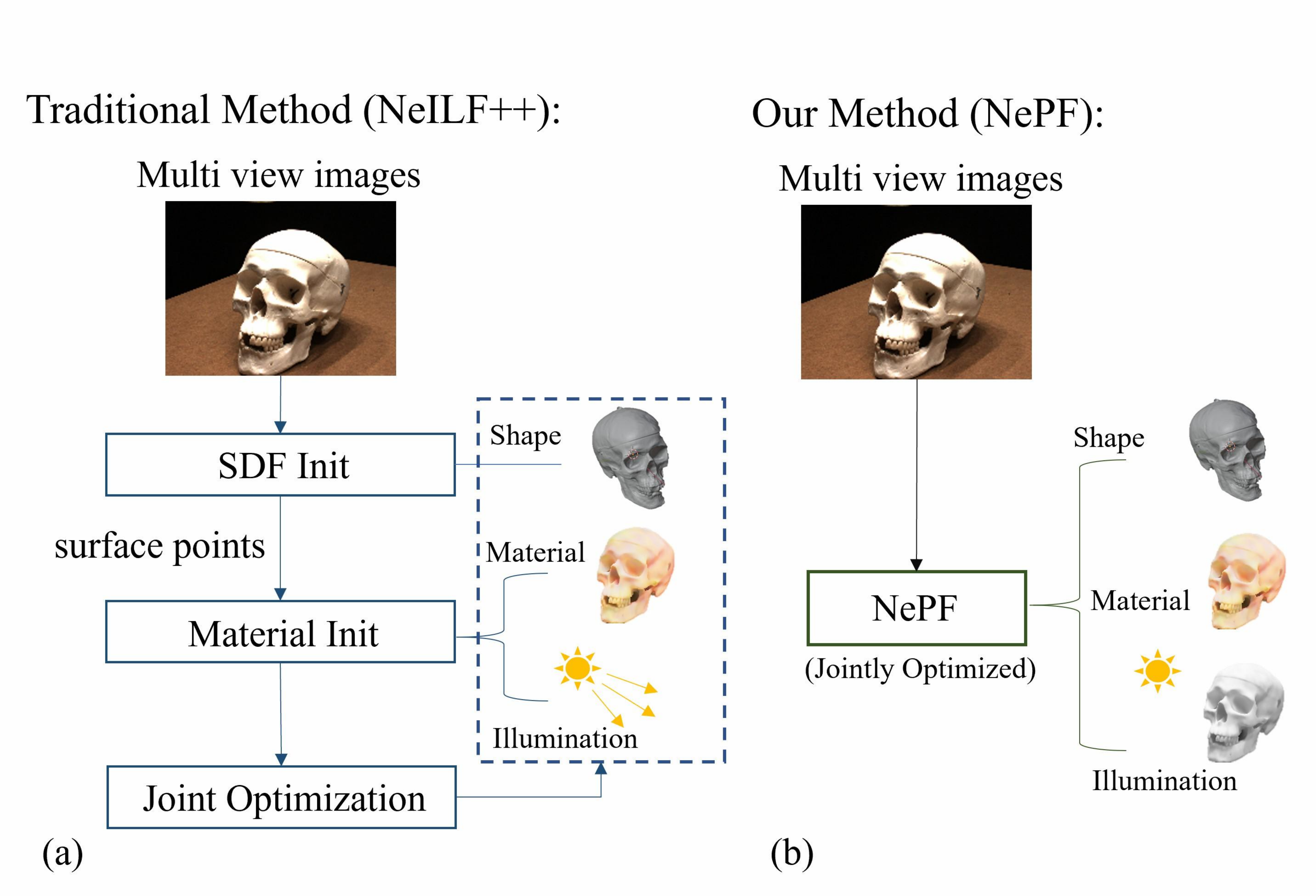}

   \caption{Comparison of the technique pipelines of the proposed single-stage NePF (b) with the SOTA three-stage  NeILF++~\cite{zhang2023neilf++} (a).} 
   \label{fig:pipe}
\end{figure}

\begin{figure}[t]
  \centering
   \includegraphics[width=1.0\linewidth]{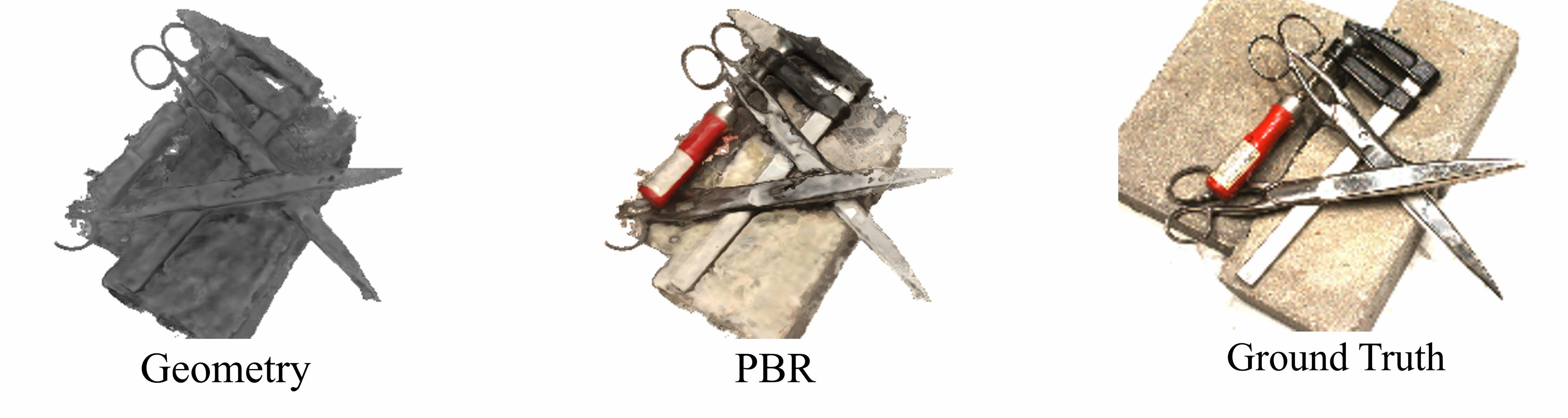}
   \caption{A typical case of accumulated error from NeRFactor~\cite{nerfactor} when training on DTU scan37~\cite{jensen2014large}. The pre-trained geometry network extracts the wrong surface and undermines the inverse rendering.} 
   \label{fig:nerfactor}
\end{figure}

First, it poses the risk of \textbf{accumulated errors} from different stages as shown in ~\cref{fig:nerfactor}, potentially leading to sub-optimal outcomes. Moreover, most previous methods~\cite{boss2021neuralpil, physg2021, nerfactor} primarily use NeRF or neural implicit representations~\cite{mildenhall2020nerf, yariv2021volume, wang2021neus} only for geometry recovery, overlooking the rich information encapsulated in the outgoing radiance field, resulting in significant \textbf{computational waste}. These have highlighted the demand for a unified, single-stage inverse rendering framework.

Appealing to this demand, we question the necessity of a muti-stage solution and introduce \textbf{\underline{Ne}}ural \textbf{\underline{P}}hoton \textbf{\underline{F}}ield, \textbf{NePF}, a single-stage framework for inverse rendering.
To address the aforementioned challenges within our method, it becomes imperative to recover the desired properties on the sampled rays together with the geometry density. In this paper, we propose to synchronize the inverse rendering with the outgoing radiance on the rays to ensure the properties are valid. We highlight the physical implications of the NeuS~\cite{wang2021neus} weight functions and view-dependent neural rendering as instrumental in merging the inverse rendering fields (where the surface property is pivotal) with the outgoing radiance field. Our experiments show that it is feasible to recover all the necessary properties by training multiple MLPs based on the same density field.

To bolster the one-stage approach and facilitate rapid volume physically-based rendering (PBR), we introduce an implicit coordinate-based illumination model for reflectance estimation that outputs both light directions and intensities along with material properties instead of the tradional environment map. To further enhance the accuracy of light directions and intensities predicted, we incorporate subsurface scattering for the diffuse component—which offers directionality in contrast to the conventional Lambertian model. While numerous established methods faltered in extracting the desired properties from intricate real-world scenes such as the DTU dataset~\cite{jensen2014large}, our technique proved resilient, generating high fidelity geometry and visual-plausible materials, all within one pass. 

The main contributions of this work can be summarized as follows.
\begin{itemize}[leftmargin=4mm, itemindent=0cm]
\item To reduce the accumulated error and computational waste in traditional neural inverse rendering, a single-stage framework is proposed which is capable of jointly optimizing all properties. To the best of our knowledge, this is the first work to address inversing rendering in a singe stage manner.
\item To accelerate the physically-based rendering in a volumetric setting, a coordinate-based illumination model is developed to directly estimates the lighting directions and intensities.
\item To regularize the predicted lighting directions and intensities, subsurface scattering is introduced for material estimation.
\end{itemize}
\section{Related Work}
\label{sec:formatting}

\subsection{Neural Implicit Surfaces for Shape Recovery}

Recovering geometry information is a critical step in inverse rendering because it deeply impacts the later material and illumination estimation. Although Neural Radiance Field (NeRF)~\cite{mildenhall2020nerf} has shown impressive results in novel view synthesis, it doesn't reconstruct geometry accurately. Some neural rendering methods enhance geometry reconstruction by adding regularizations to the NeRF density field~\cite{nerfactor, refnerf}, by implementing occupancy field~\cite{Oechsle2021ICCV, s3nerf, occnetwork, dvr}, or by trainning a SDF function via volume rendering~\cite{yariv2021volume,wang2021neus,petneus, neuralwarp,li2023neuralangelo,mvsdf}. Notably, NeuS (NeuS)~\cite{wang2021neus} has been particularly effective in reconstructing scene geometry using volume rendering of an SDF density field. 

\begin{figure}[t]
  \centering
   \includegraphics[width=1.0\linewidth]{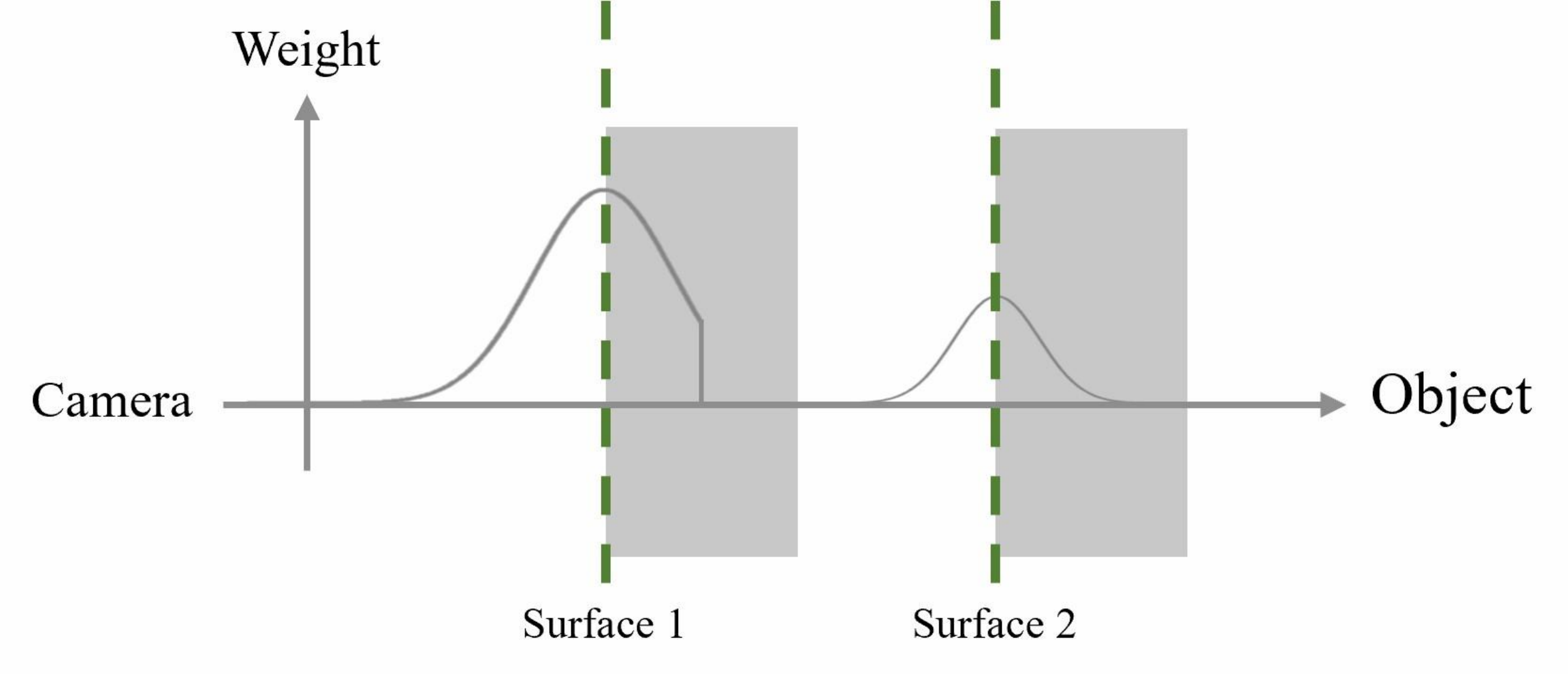}
   \caption{A simple demonstration of NeuS weight function. The weight reaches its maximum at the surface near the camera.} 
   \label{fig:weight}
\end{figure}

One key innovation in NeuS~\cite{wang2021neus} is its unique weight function, which is unbiased and occlusion-aware. For surfaces closer to the camera, it hits its highest value, making it more significant in the final rendering as presented in \cref{fig:weight}. This feature enables NeuS~\cite{wang2021neus} to create realistic 3D geometry from multi-view images.
The physical implication behind this weight function led us to integrate geometry recovery into our inverse rendering in a single-stage manner. We can locate the surface property according to the density field by simply querying the max weight. Hence, we choose NeuS~\cite{wang2021neus} as our baseline for geometry recovery. The shape is represented as:
\begin{equation}
 \mathcal{S}: \left\{ x\in \mathbb{R}^3 | f(x)=0\right\} .
\end{equation}

\subsection{Illumination and Material Estimation}
Another aspect of inverse rendering is to model the illumination effect to estimate the surface materials. Several previous methods~\cite{bi2020nrf, nerv2021, boss2021neuralpil} need prior knowledge or pre-trained model of illumination for material estimation. Others assume the illumination to be point/colocated flash lights~\cite{drv,bi2020deep,Schmitt2020CVPR,XiaDPT16}. Recently, several methods have proved that it is effective to model the illumination by using environmental maps~\cite{boss2021nerd,physg2021, nerfactor,munkberg2023extracting}. These recent methods, due to their inherent computational complexity, restrain their application only for surface rendering, making it essential to make surface locating a separate stage. Also, the ill-posed nature of the inverse rendering makes these methods hard to distinguish the environment color and the original color, which leads to intuitively incorrect results.

Recently, Neural Incident Light Field (NeILF)~\cite{neilf} presents a coordinate-based illumination model and achieves satisfying results. However, the neural light field is still separate from the outgoing radiance field, which brings complexity and incoherence. In this work, we come up with a new illumination model that shares the same density field with the outgoing radiance field. This method, by directly regressing the light directions and intensities, enables fast forward rendering in a volumetric setting. 

Provided the illumination information in the scene, the material can be estimated by comparing the ground truth rgb with the color of physically-based rendering (PBR) by Bidirectional Reflectance Distribution Function (BRDF) or Bidirectional Scattering Distribution Function (BSDF). Traditional inverse rendering models use BRDF for simplicity. In our work, to better regularize the light directions, we introduce Disney Principled BSDF, which replaces the Lambertian model in BRDF with subsurface scattering, on account of it provides directionality in diffuse part.
\subsection{Neural Inverse Rendering}

Recovering the geometry, material, and illumination from images is an ill-posed task. Recent approaches~\cite{boss2021neuralpil, nerfactor, physg2021, zhang2023neilf++} show that it's possible to recover all these factors using only the true colors in an image as a guide using neural implicit representations.

However, these methods often involve a multi-stage strategy. They first estimate the geometry, then locate the surface coordinates, and finally use these coordinates in further models. PhySG~\cite{physg2021} and NeRFactor~\cite{nerfactor} need an initialization stage for shape reconstruction to extract the surface point, and then perform the rest of the inverse rendering. The best of the SOTA methods: NeILF++~\cite{zhang2023neilf++}, successfully achieves the joint optmization of shape, material and illumiantion via the inter reflections between surfaces, but it still requires initializations of geometry and material before the final optimization. In our work, we aim to unify these steps. By focusing on the geometric implication of the weight function produced by NeuS~\cite{wang2021neus}, we propose to further optimize the outgoing radiance field together with the inverse rendering module all in one pass.

\section{Methods}
\label{sec:methods}

Our objective is to extract geometry, material, and illumination for an object in a single-stage manner using a collection of multi-view images taken under unknown lighting conditions. We operate under the assumption that each coordinate in space is predominantly influenced by one major light source, which complies with most indoor real-world scenrarios.

\subsection{NeuS Weight Function}
\label{sec:neus}
The efficacy of NeuS~\cite{wang2021neus} in geometry recovery is rooted in its distinctive weight function, which is unbiased and occlusion-aware. This weight function reaches its highest value near the camera surface and local peaks at subsequent surfaces, as shown in \cref{fig:weight}.

As a result, it retains rich geometric information from the scene. By querying the max weight along the ray, we can easily get the surface PBR result. The weight function is defined as follows:
\begin{align}
W(t) = \prod^{}_{j<t}(1-\alpha(j))\cdot \alpha(t) ,\\
 \alpha(t) = 1 - exp(-\int_{t_i}^{t_{i+1}}p(t)dt) ,
\end{align}
where $p$ refers to the opaque density:
\begin{equation}
p(t) = max\left(\dfrac{-\dfrac{d\phi_{s}}{dt}(f(x(t)))}{\phi_{s}(f(x(t)))},0\right) ,
\end{equation}
where $\phi_s$ is the logistic density function, $f(x)$ represents the signed distance function.
\subsection{Single-Stage Inverse Rendering}

\begin{figure}[t]
  \centering
   \includegraphics[width=1.0\linewidth]{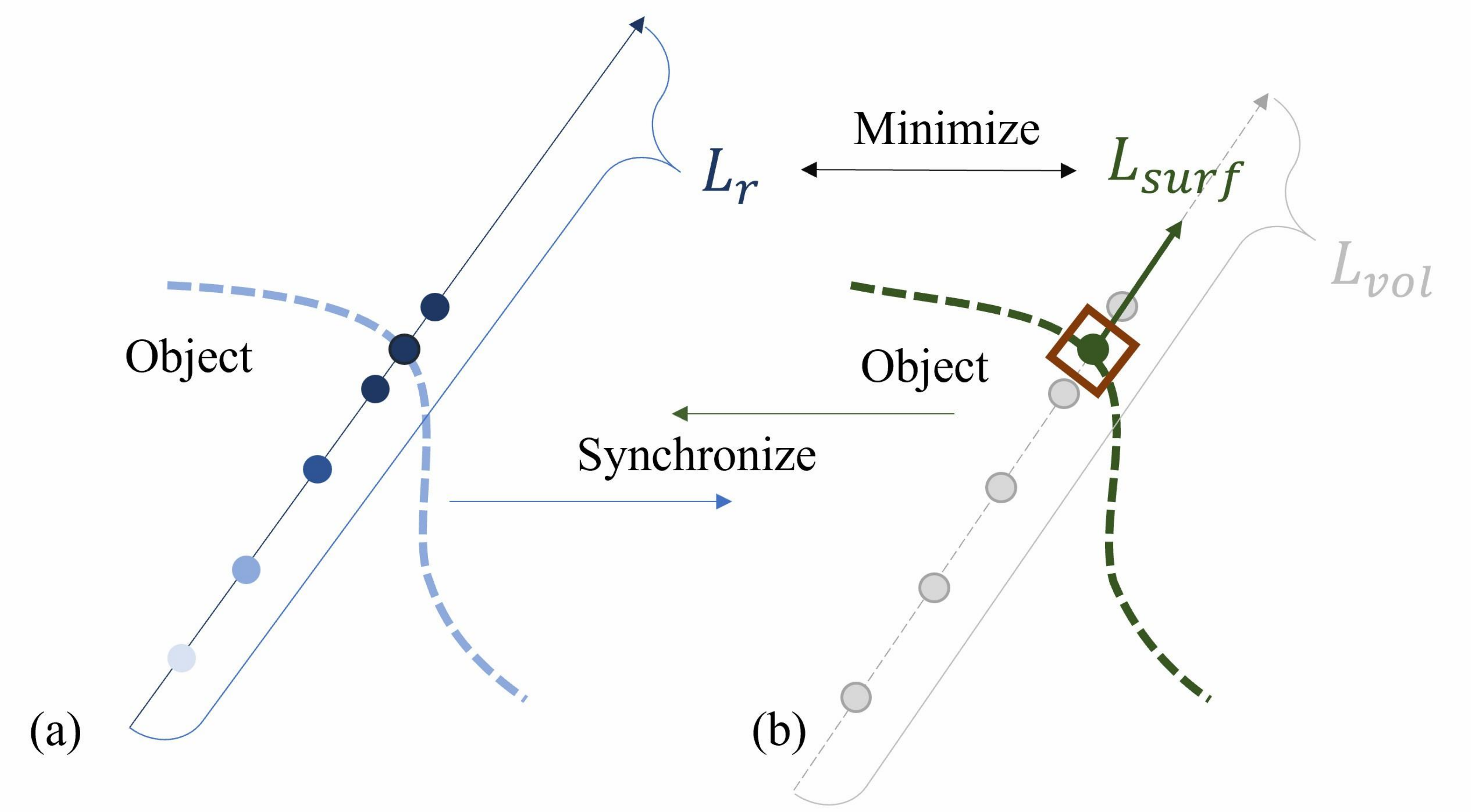}

   \caption{Comparison of the volume rendering and the physical-based rendering. (a) shows that, for the outgoing radiance field, the volumetric radiance is taken into consideration for the final rendering result, while for physical-based rendering (b), only the surface points are considered. The synchronization of this contradicted focus is achieved through supervising the physical-based result (b) under the rendering outcome of (a).} 
   \label{fig:synchron}
\end{figure}

\indent  While inverse rendering, commonly based on the hard surface assumption, only accounted for surfaces, neural radiance rendering emphasizes volumetric field information. Therefore, the contradicted focal points of the two tasks have kept researchers away from further utilizing the neural radiance field (\cref{fig:synchron}).
 \\
\indent As a result, traditional methods~\cite{boss2021neuralpil, physg2021, nerfactor, zhang2023neilf++} usually recover the geometry, or initialize the shape first, then locate the surface points, and finally bridge geometry accuracy with subsequent material and illumination estimated on these surface points (\cref{fig:pipe}). To unify these steps in one pass, we will have to focus on the volumetric space and recover all properties within the same density field. We will introduce our framework from the direct solution of a single-stage inverse rendering .\\
\textbf{Direct solution: }A direct solution is to train a volumetric inverse rendering field using ground truth color as supervision, and extract the surface property by querying the max weight along the ray according to \cref{fig:weight}. However, it produces inaccurate geometries because of the intricacies of volume rendering and the broad solution space of the inverse rendering, as shown in \cref{fig:supervision}. \\
\textbf{NePF solution: }To make sure the geometry property is valid, it is essential to train the density field of which the view-dependent outgoing radiance is supervised by the ground truth color. And if we can regress the inverse rendering properties based on the same density field together, then the problems we encounter in direct baseline will be solved. However, several challenges remain.
\begin{itemize}
\item How do we regularize the surface properties of inverse rendering in the volume space?
\item How do we ensure the inverse rendering field interprets the scene volumetrically in alignment with the outgoing radiance field?
\end{itemize}

\begin{figure}[t]
\centering
\includegraphics[width=1\linewidth]{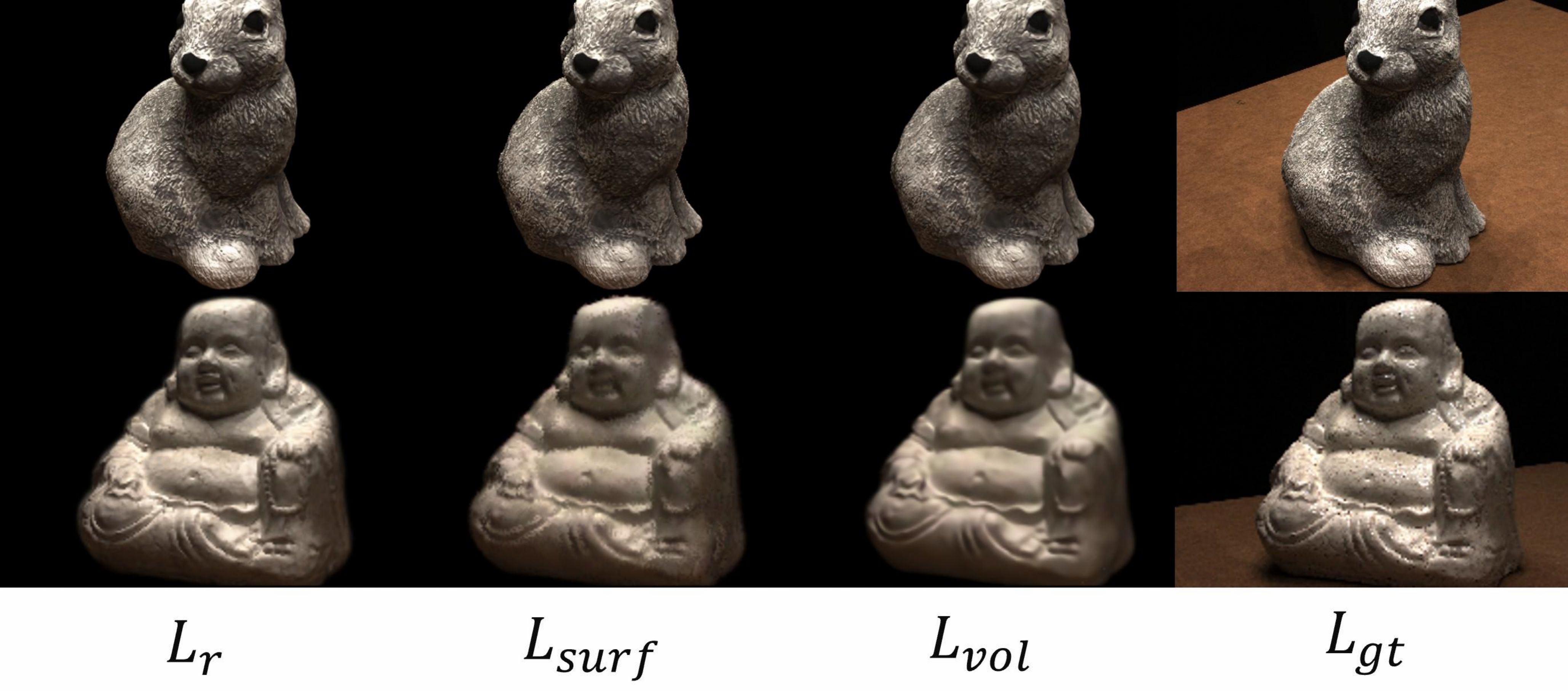}
\caption{Demonstration of different fields' synchronizations. The $L_r$, $L_{surf}$, $L_{vol}$ represent the rendering of the outgoing radiance field, surface by querying max weight and the inverse rendering field. The figure shows that the synchronization of two fields is successfully built, while the surface property is valid.}
\label{fig:synch2}
\end{figure}

Addressing these challenges requires a connection to relate the two distinct fields, enabling synchronized interpretations on the object. This connection can be found in the physical implications of the rendering outcome by the neural radiance field. The view-dependent characteristic of the outgoing radiance field physically arises from spatially-varying reflectance on surfaces. As such, the outgoing radiance field can serve as a \textbf{pseudo ground truth} for the surface property in the inverse rendering field during training. By matching the color $L_{surf}$ of the surface points $x_{maxweight}$ with the rendering results $L_r$ of the outgoing radiance, we ensure the outcome of inverse rendering on the surface is as expected. Also, the selections of these surface points, determined by max weights based on the density field, impose an additional focus on the accurate surfaces that penalize unwanted shape distortions, which is common in the direct baseline. Moreover, by regularizing the surface color, the inverse rendering field is also adjusted, given that the MLPs representing various features are continuous in 3D space. Hence, we synchronize the two distinct neural fields by minimizing the difference between the rendering of view-dependent radiance $L_r$ and the surface physically-based rendering $L_{surf}$ as shown in \cref{fig:synch2}, and we can now train the inverse rendering field concurrently with the outgoing radiance field.\\
\textbf{Shape Refinement: }The geometry extracted solely under the supervision of outgoing radiance is not accurate enough. Since the view-dependent outgoing radiance does not consider specular effects, it could lead to potential errors in shape recovery. Moreover, we find that current neural implicit surfaces methods~\cite{wang2021neus, yariv2021volume} tend to overfit the textures as shape. Therefore, we propose to further regularize the density field by trainning the inverse rendering field $L_{vol}$ under the supervision of ground truth color $L_{gt}$. Our experiments in \cref{sec:abl} prove that this can help improve the quality of geometry recovery.
\\
\textbf{Loss function: }We calculate three l1 losses for each part, where $\mathcal{L}_r = \|L_r - L_{gt}\|$,  $\mathcal{L}_{surf} = \|L_{surf} - L_r\|$,  $\mathcal{L}_{vol} = \|L_{vol} - L_{gt}\|$. Our overall loss function for color can be defined as follows: 
\begin{equation}
    \mathcal{L} = \mathcal{L}_r + \lambda _1 * \mathcal{L}_{surf}  + \lambda _2 * \mathcal{L}_{vol} .
\end{equation}
We empirically set $\lambda_1$ = 0.0003 and $\lambda_2$ = 0.0001  in our experiments.\\
\textbf{Surface Approaching: }It's important to note that in actual applications, the structure might not always sample the exact surface points, especially at the early stages of training. However, we've observed that this doesn't compromise the overall performance, with the network progressively focusing on points near the surface due to the importance-sampling mechanism. Still, to help the network converge faster to the surface points, we multiply an additional $(1-weight_{max})$ to the $\mathcal{L}_{surf}$. 
\begin{figure*}
  \centering
   \includegraphics[width=0.95\linewidth]{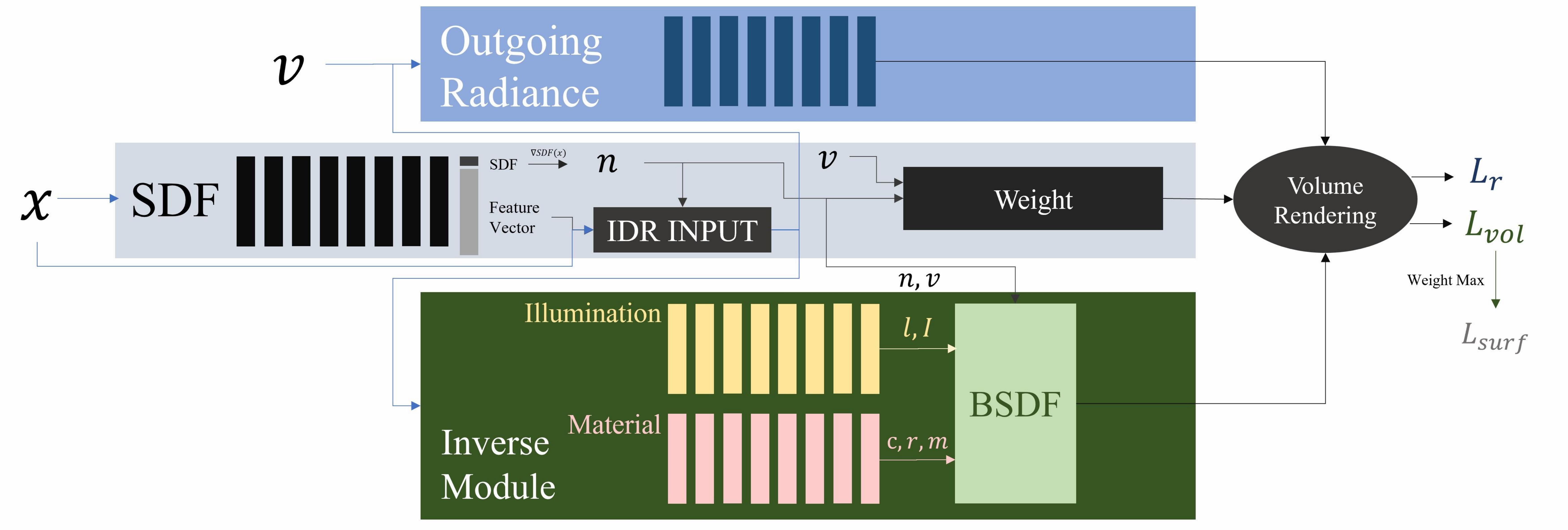}
   \caption{The overall structure of NePF. The SDF network first takes the coordinates $x$ as input and then outputs the SDF value along with a 256-dimension feature vector. We calculate the gradient of SDF to get the surface normals. Then, we concatenate the coordinates with normals and feature vectors as the input for NePF. The illumination MLP $\mathcal{P}$ and the Material MLP $\mathcal{M}$ return the information needed for BSDF estimation. We perform two volume renderings based on the weight function provided by SDF along the ray. We get the PBR outcome on the surface by querying the max weight. The outgoing radiance field remains the same structure as in NeuS~\cite{wang2021neus} with sine activations as in SIREN~\cite{sitzmann2019siren}. More details can be found in supplementary materials.} 
   \label{fig:SSR}
\end{figure*}
\subsection{Coordinate-based Illumination}
\indent We use a straighforward method for illumination. It mirrors the IDR~\cite{yariv2020idr} input,where normals and feature vectors are concatenated with coordinates. The network produces a unit-length vector $l$ calculated by two angles $\rho$, $\phi$, representing light directions and a corresponding radiance value $I$. The network for illumination is outlined as follows: 
\begin{equation}
    \mathcal{P}: \left\{ x,n,feature \right\} \rightarrow \left\{\rho,\phi,I \right\} .
\end{equation}
For clarity, this MLP's output isn't related to photon mapping, even though our inspiration came from it. We adopt this name to provide readers with an intuitive grasp. As this illumination is coordinate-driven, its information can be directly applied for efficient volume physically-based rendering.
\subsection{BSDF Modeling and Material Estimation: }
\indent To ensure a valid estimation of lighting direction, we incorporate subsurface scattering in place of the Lambertian model. The former offers directional regularization. We utilize the Disney Principled BSDF within our method. The diffuse component is defined as follows:
\begin{equation}
    f_{d}=\dfrac{c}{\pi }\left( 1-\dfrac{f_{l}}{2}\right) \cdot \left( 1-\dfrac{f_{v}}{2}\right) +f_{retro} .
\end{equation}
The $f_{retro}$ is retro reflection term:
\begin{equation}
        f_{retro}=\dfrac{c}{\pi}\cdot R\cdot \left[ f_{l}+f_{v}+f_{l}\cdot f_{v}\cdot \left( R-1\right) \right] ,
\end{equation}
where $f_{l}$, $f_{v}$ are defined by
\begin{align}
    f_l=(1-n\cdot l)^5   &\ , \ f_v=(1 - n \cdot v)^5 ,
\end{align}
and $R$ is calculated by 
\begin{align}
 R= 2 \cdot r &\cdot (h \cdot v)^2.
\end{align}
The specular term is computed as follows:
\begin{equation}
    f_{s} = \dfrac{F(v, h;c,m) \cdot G(l,v;h,r) \cdot D(h,r)}{4 \cdot (n\cdot v) \cdot (n\cdot l)} ,
\end{equation}
where $F, G, D$ denote the Fresnel term, geometry term, and normal distribution term, respectively. More implementation details can be found in supplementary materials.

In the above equations, we use $c$ to represent albedo, $r,m$ to represent roughness and metallic as material attributes, $l,v,n$ represent light direction, view direction and normals respectively. $h$ is the half vector of $l$ and $v$.

The same input structure as the illumination part is followed by the material part, producing a 5D vector encompassing roughness, metallic values that both range from [0, 1] and albedo value that ranges from [0, 1]$^3$. The material network is defined as follows:
\begin{equation}
    \mathcal{M}:\left\{ x,n,feature \right\} \rightarrow \left\{c, r, m\right\} .
\end{equation}
Feeding these parameters into the BSDF model enables us to conduct physically-based renderings.
 \begin{figure*}[t]
\centering
\includegraphics[width=1\linewidth]{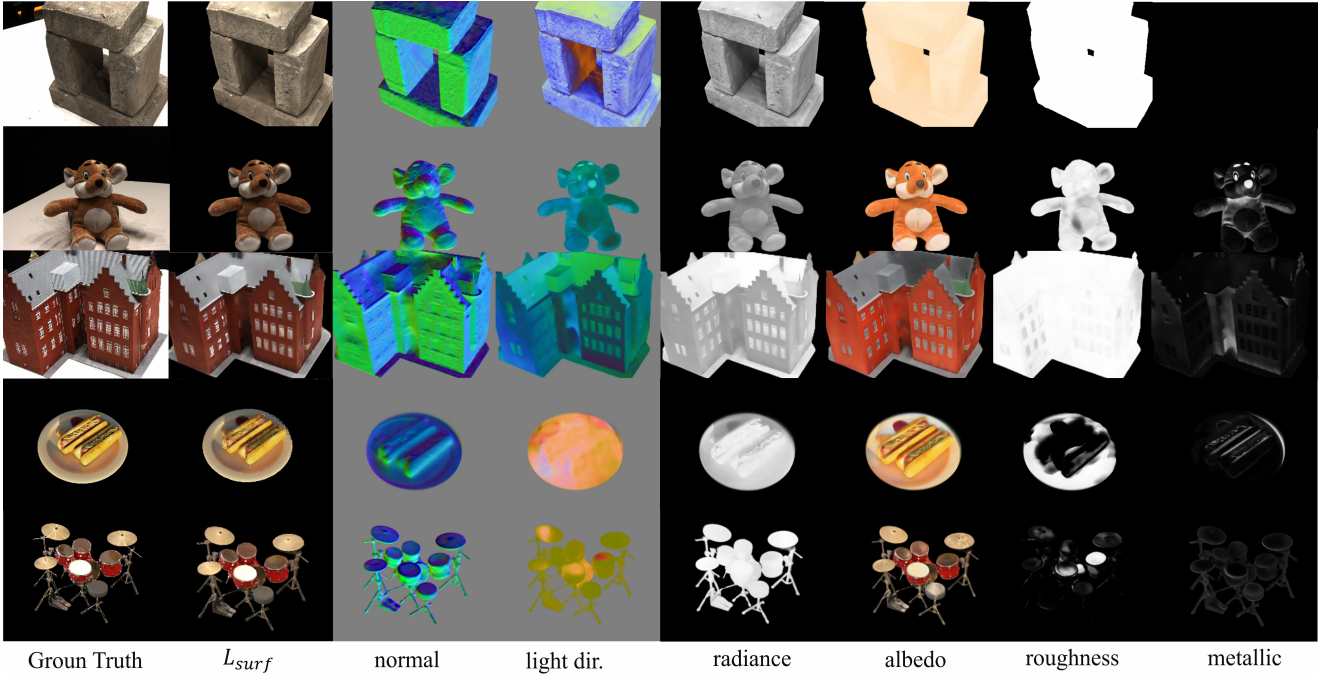}
\caption{Qualitative results on DTU~\cite{jensen2014large} and NeRF synthetic datasets~\cite{mildenhall2020nerf}. The results show that the proposed method successfully recovered the high-fidelity geometry and accurate novel view synthesis.}
\label{fig:all}
\end{figure*}
\subsection{Rendering}

In our framework, we concurrently execute two volume renderings~\cite{Kajiya1986TheRE}. The first originates from outgoing radiance field, while the other, rooted in our Neural Photon Field baseline, considers the BSDF for subsurface scattering and specular effects:
\begin{align}
    L_r &=\int _{\Omega }W\left( t\right) \cdot L\left( x\left( t\right) ,v \right) dt ,\\
    L_{vol} &=\int _{\Omega }W\left( t\right) \cdot f_{BSDF}(l,v,h,n,c,m,r) \cdot I( x\left( t\right)) dt .
\end{align}
The weight funcion $W(t)$ is defined in \cref{sec:neus}. Our entire approach is illustrated in \cref{fig:SSR}.
In conclusion, our single stage inverse rendering framework NePF can be briefly described as:
\begin{itemize}
\item Render the results of outgoing radiance field $L_{r}$ and inverse rendering field $L_{vol}$, both share the same density field in space.
\item Locate the surface PBR result $L_{surf}$ by querying the max weight.
\item Minimize the difference between the $L_{surf}$ with $L_{r}$ to synchronize two fields.(\cref{fig:synchron})
\item Supervise $L_{r}$ and $L_{vol}$ under the ground truth color to adjust the density field.
\end{itemize}

\section{Experiments}

We test our method on DTU datasets~\cite{jensen2014large} and NeRF synthetic datasets~\cite{mildenhall2020nerf}. We further evaluate the significance of some of our key innovations in ablation studies. Since most of the previous methods promise a satisfying result over the synthetic datasets, to signify our improvements, we use the challenging DTU~\cite{jensen2014large} datasets mainly for quantitative and qualitative comparison.

\subsection{Experimental setting}
\indent \textbf{Baseline:} We compare our model with several current methods on neural inverse rendering, PhySG, Neural PIL, NeRFactor, and NeILF++~\cite{physg2021,boss2021neuralpil,nerfactor,zhang2023neilf++}. We further test the geometry reconstruction with NeILF++~\cite{zhang2023neilf++} and NeRF~\cite{mildenhall2020nerf} since several SOTA methods~\cite{nerfactor, boss2021neuralpil} use NeRF for geometry recovery. \\
\textbf{Implementation details:} We sample 512 rays per epoch. We train our network with 250000 iterations for around 20 hours on a single Nvidia V100 GPU. More details of the training scheme are in supplementary materials.

\subsection{PBR Novel View and Shape Reconstruction}
\indent We extract the meshes at where SDF=0 and generate the novel view of PBR on the surface by locating the global maxima of weights along the ray. \\
\begin{table*}[t]
\caption{Quantitative results obtained by the proposed method and the comparison methods on DTU scans~\cite{jensen2014large}. The best one is marked as red, while the second and third are marked as orange and yellow respectively, in novel view synthesis chart. Our framework successfully ties the best of SOTA methods, NeILF++~\cite{zhang2023neilf++} in terms of PBR novel view synthesis. Although our method cannot recover high quality geometry compare to NeILF++~\cite{zhang2023neilf++} on average, in certain cases, we show potential to outperform the best SOTA method. The results of other methods are from NeILF++~\cite{zhang2023neilf++}.}
  \centering
  \small 
  \setlength{\tabcolsep}{2pt} 
  \begin{tabular}{@{}l|@{}p{2cm}|c c c c c c c c c c c c c c c |c}
    \toprule
    Task & \textbf{Methods} & 24 & 37 & 40 & 55 & 63 & 65 & 69 & 83 & 97 & 105 & 106 & 110 & 114 & 118 & 122 & \textbf{Mean} \\
    \midrule
    & PhySG~\cite{physg2021} &15.11 & 17.31& 17.38&20.65 & 18.71& 18.89& 18.47& 18.08& 21.98& 20.67& 18.75& 17.55& 21.20& 18.78& 23.16&19.11 \\
    Novel & Neural PIL~\cite{boss2021neuralpil} & 19.51& 19.88& 20.67& 19.12& 21.01& \cellcolor{yellow!75}23.70& 18.94& 17.05& 20.54& 19.67& 18.20& 17.75& \cellcolor{yellow!75}21.38&21.69 & - & 19.94\\
    View & NeRFactor~\cite{nerfactor} &\cellcolor{yellow!75}21.91&\cellcolor{yellow!75}20.45&\cellcolor{yellow!75}23.24&\cellcolor{yellow!75}23.33&\cellcolor{yellow!75}26.86&22.70&\cellcolor{yellow!75}24.71&\cellcolor{yellow!75}27.59&\cellcolor{yellow!75}22.56&\cellcolor{yellow!75}25.08&\cellcolor{yellow!75}26.30&\cellcolor{yellow!75}25.14&21.35&\cellcolor{yellow!75}26.44&\cellcolor{yellow!75}26.53&\cellcolor{yellow!75}24.28 \\
    (PSNR$\uparrow$)& NeILF++~\cite{zhang2023neilf++} &\cellcolor{orange!75}24.17 &\cellcolor{red!75} 24.60& \cellcolor{orange!75}26.40& \cellcolor{red!75}27.24&\cellcolor{red!75} 29.85& \cellcolor{red!75}28.16& \cellcolor{orange!75}27.39&\cellcolor{orange!75}29.82 &\cellcolor{orange!75}25.50 & \cellcolor{orange!75}28.19&\cellcolor{red!75} 31.84& \cellcolor{red!75}30.20&\cellcolor{red!75}27.71 &\cellcolor{red!75} 30.87&\cellcolor{red!75} 33.62&\cellcolor{red!75}28.37 \\
     & Ours &\cellcolor{red!75}25.66&\cellcolor{orange!75}21.88&\cellcolor{red!75}28.34&\cellcolor{orange!75}27.14&\cellcolor{orange!75}29.29&\cellcolor{orange!75}27.92&\cellcolor{red!75}28.94&\cellcolor{red!75}32.00 &\cellcolor{red!75}26.35&\cellcolor{red!75}28.27&\cellcolor{orange!75}30.33& \cellcolor{orange!75}27.64&\cellcolor{orange!75}26.49&\cellcolor{orange!75}28.82&\cellcolor{orange!75}31.08& \cellcolor{orange!75} 28.01\\
    \midrule
    Geometry&  NeRF~\cite{mildenhall2020nerf} & 1.920 & 1.730& 1.920&0.800 &3.410 &1.390 &1.510 &5.440 &2.040 &1.100 &1.010 &2.880 & 0.910&1.000 &0.790 &1.857 \\
   (Chamfer & NeILF++~\cite{zhang2023neilf++} &1.303 & 1.911& 0.954&0.595 & 1.271& 0.881& 0.860&1.529 &1.242 &1.125 &0.723 & 2.096& 0.488&0.838 & 0.654& 1.098 \\
    Distance$\downarrow$)&Ours &1.716&3.474 &\textbf{0.724}& \textbf{0.413}& 2.222 &1.064 &1.801 &1.807&2.078&1.132 & \textbf{0.600} &2.432 &0.760&1.099&0.740 & 1.472 \\
    \midrule   
    \bottomrule
  \end{tabular}

  \label{tab:quantitive_all}
\end{table*}
Since inverse rendering is an ill-posed problem, we compared our model with other SOTA methods in terms of PBR novel view synthesis for quantitative comparison. We further evaluate the geometry accuracy based on the Chamfer Distance. We also provide the material estimation as well as the illumination estimation in ~\cref{fig:all}.
As ~\cref{tab:quantitive_all} shows, in novel view synthesis, our methods accomplish approximately the same, in some cases even better results comparing to NeILF++~\cite{zhang2023neilf++}, the best SOTA method which implements three stages for inverse rendering, in a single stage manner. Also, in most of the test cases, our method is able to recover high-fidelity geometry. These data prove the efficacy of our single-stage method. \\
\subsection{Ablation Studies}
\label{sec:abl}
\begin{figure}[h]
\centering
\includegraphics[width=1.0\linewidth]{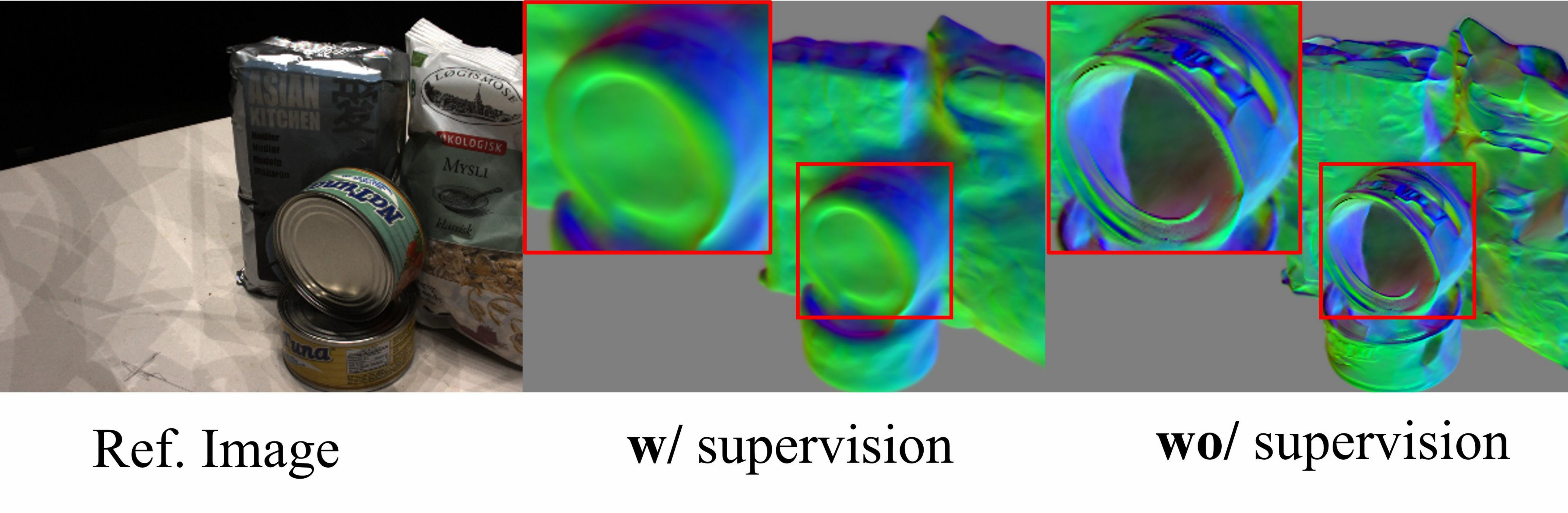}
\caption{Qualitative study on outgoing radiance supervision. The geometry recovered without the outgoing radiance supervision is unreliable.}
\label{fig:supervision}
\end{figure}

\begin{figure}[h]
\centering
\includegraphics[width=1.0\linewidth]{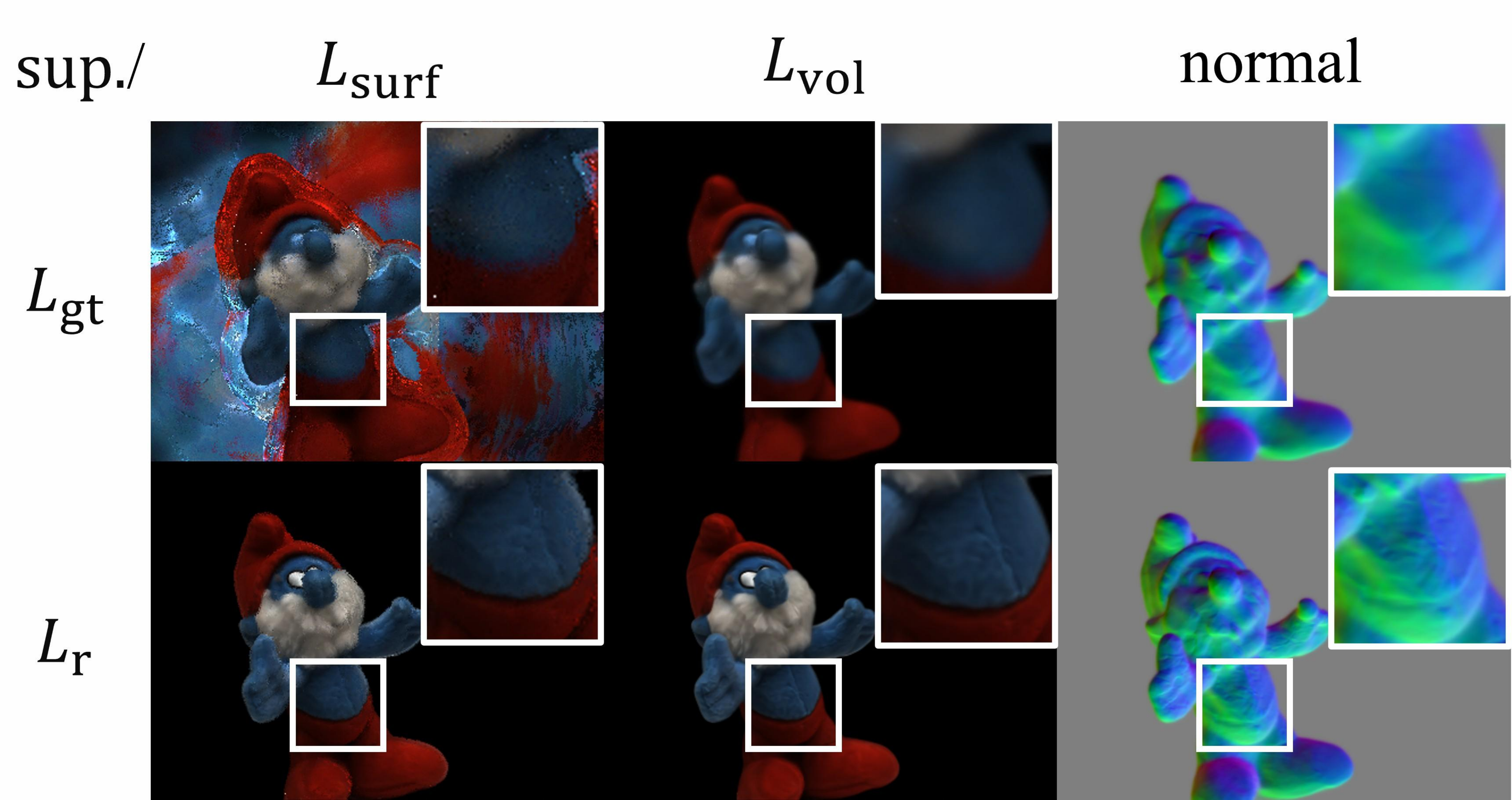}
\caption{Qualitative study on two ways of supervision on $L_{surf}$ on DTU scan83~\cite{jensen2014large}. Despite the apparent overfitting in the background, when supervised under the ground truth, the overall novel view PBR quality is significantly lower, as well as the geometry recovery.}
\label{fig:gt_supervised}
\end{figure}

\indent\textbf{Outgoing Radiance Field Supervision:} As the essence of our method, we explore the necessity of outgoing radiance field supervision in \cref{fig:supervision}
As the figure shows, in the absence of outgoing radiance supervision, the network wrongly predicts the geometry in the square due to the immense solution set of inverse rendering.\\
\textbf{The Color Loss: }The key to synchronizing the inverse rendering field with the outgoing radiance field is to supervise the surface PBR results $L_{surf}$ with the volume rendering of outgoing radiance $L_{r}$. To explain why we have to supervise the surface PBR results with the $L_{r}$ during trainning instead of the ground truth RGB $L_{gb}$, we test the model with the surface PBR $L_{surf}$ supervised under different pseudo ground truth in \cref{fig:gt_supervised} and \cref{tab:gt_supervised}. 
\begin{table}[t]
\caption{Quantitative study on two ways of supervision on $L_{surf}$ on DTU scan83~\cite{jensen2014large}. In terms of PBR novel view synthesis and geometry recovery, supervised under the ground truth color $L_{gt}$ leads to worse results.}
  \centering
  \begin{tabular}{@{}l c c c}
    \toprule
    sup. methods & Novel View (PSNR)$\uparrow$ &Geometry (CD)$\downarrow$ \\
    \midrule
    $L_{gt}$ &27.9577 & 1.8913\\
     $L_{r}$ &\textbf{31.9960} & \textbf{1.8074}\\
    \bottomrule
  \end{tabular}
  
  \label{tab:gt_supervised}
\end{table}
As presented in both \cref{fig:gt_supervised} and ~\cref{tab:gt_supervised}, supervising the surface loss $L_{surf}$ with the radiance field loss $L_{r}$ yields improved outcome. This enhancement can be attributed to the fact that the ground truth color encompasses a broader spectrum of high-frequency details when compares to the radiance field’s volume rendering, especially during the early phases of training. Given that the density field's regression is predominantly dependent on the radiance field's volume rendering results, supervising the $L_{surf}$ under the $L_{gt}$ leads to a larger overall loss that will affect the shape recovery, as well as the later PBR, then supervising under the $L_{r}$.

\begin{figure}[t]
\centering
\includegraphics[width=1.0\linewidth]{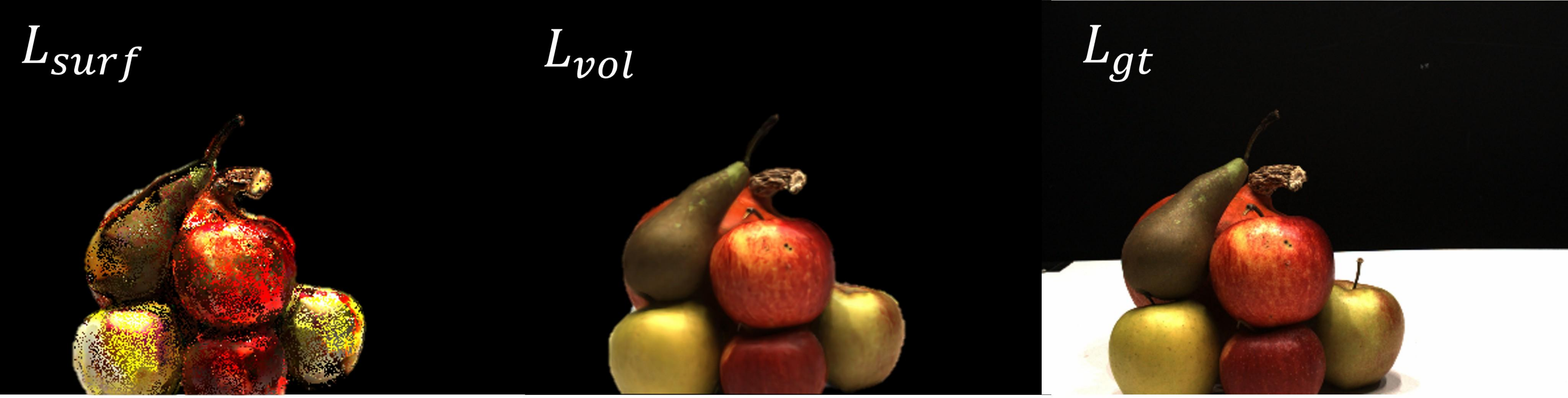}
\caption{Qualitative study on surface loss $\mathcal{L}_{surf}$. As shown in the figure, without the surface loss, the $L_{surf}$ remains noisy while $L_{vol}$ already converges.}
\label{fig:wosurf}
\end{figure}

\begin{figure}[h]
\centering
\includegraphics[width=1.0\linewidth]{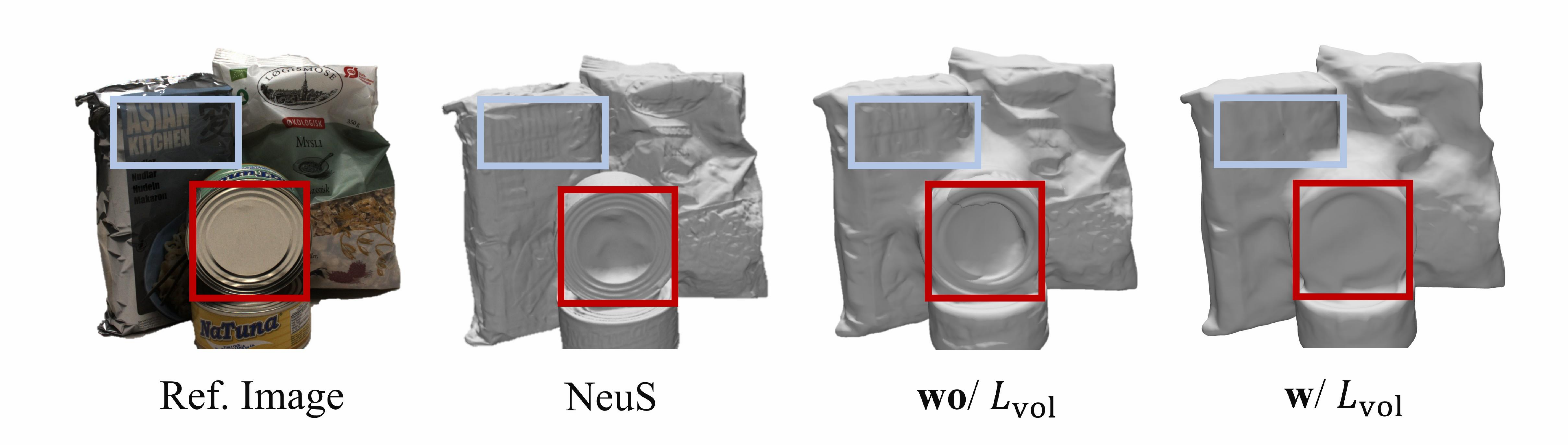}
\caption{Qualitative study on the effect of volume loss $\mathcal{L}_{vol}$. Our method is able to distinguish the textures and specular effects from shape.}
\label{fig:wovol}
\end{figure}

Also, to better understand the role of each component in our color loss, we test the scenarios where the surface loss $\mathcal{L}_{surf}$ and the volume loss $\mathcal{L}_{vol}$ are not used in \cref{fig:wosurf}, \cref{tab:wovol} and \cref{fig:wovol}.
\begin{table}[h]
  \caption{Quantitative study on the effect of volume loss $\mathcal{L}_{vol}$ on the DTU datasets~\cite{jensen2014large}. As the table demonstrates, without the volume loss, the chamfer distance of the geometry recovery is significantly higher.}
  \centering
  \begin{tabular}{@{}l c c c}
    \toprule
    Methods &scan40&scan63&scan97 \\
    \midrule
    w/ $\mathcal{L}_{vol}$ &\textbf{0.724} & \textbf{2.222}& \textbf{2.078}\\
     wo/ $\mathcal{L}_{vol}$ &0.875 & 2.909& 2.577\\
    \bottomrule
  \end{tabular}

  \label{tab:wovol}
\end{table}

\begin{figure}[t]
\centering
\includegraphics[width=1.0\linewidth]{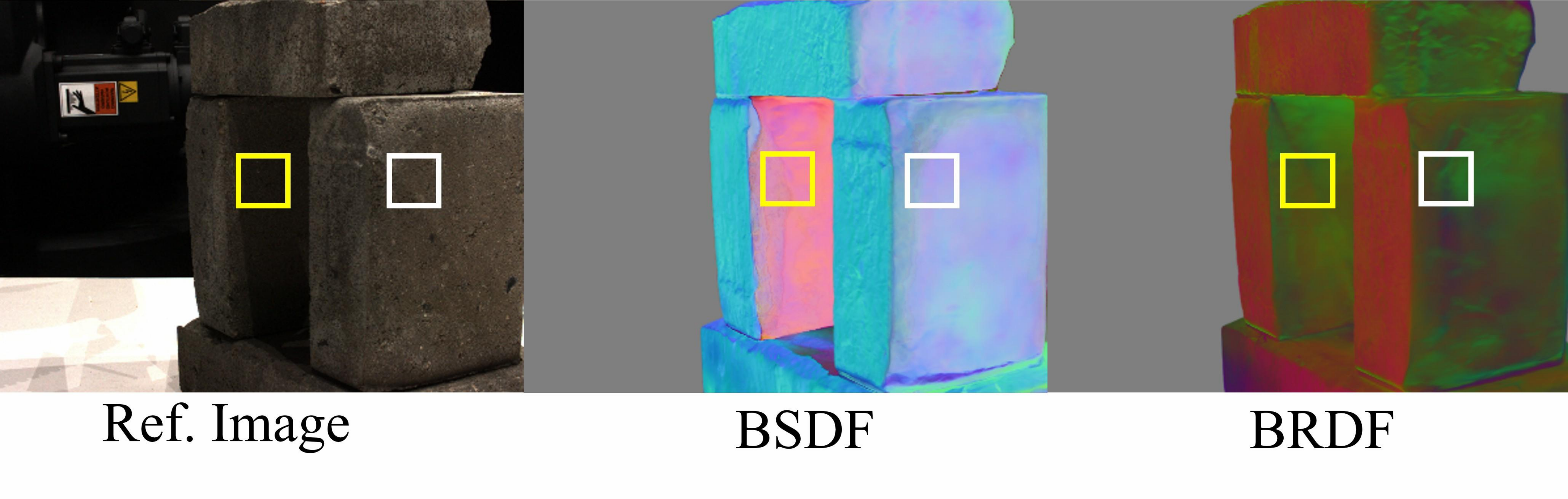}
\caption{Qualitative study on subsurface scattering. The images show the light directions under the BSDF and BRDF models. The white square and the yellow square are under different illumination contexts due to the occlusion. While the BSDF model can tell such differences, the BRDF fails to do so.}
\label{fig:sss}
\end{figure}

\begin{figure}[t]
\centering
\includegraphics[width=1.0\linewidth]{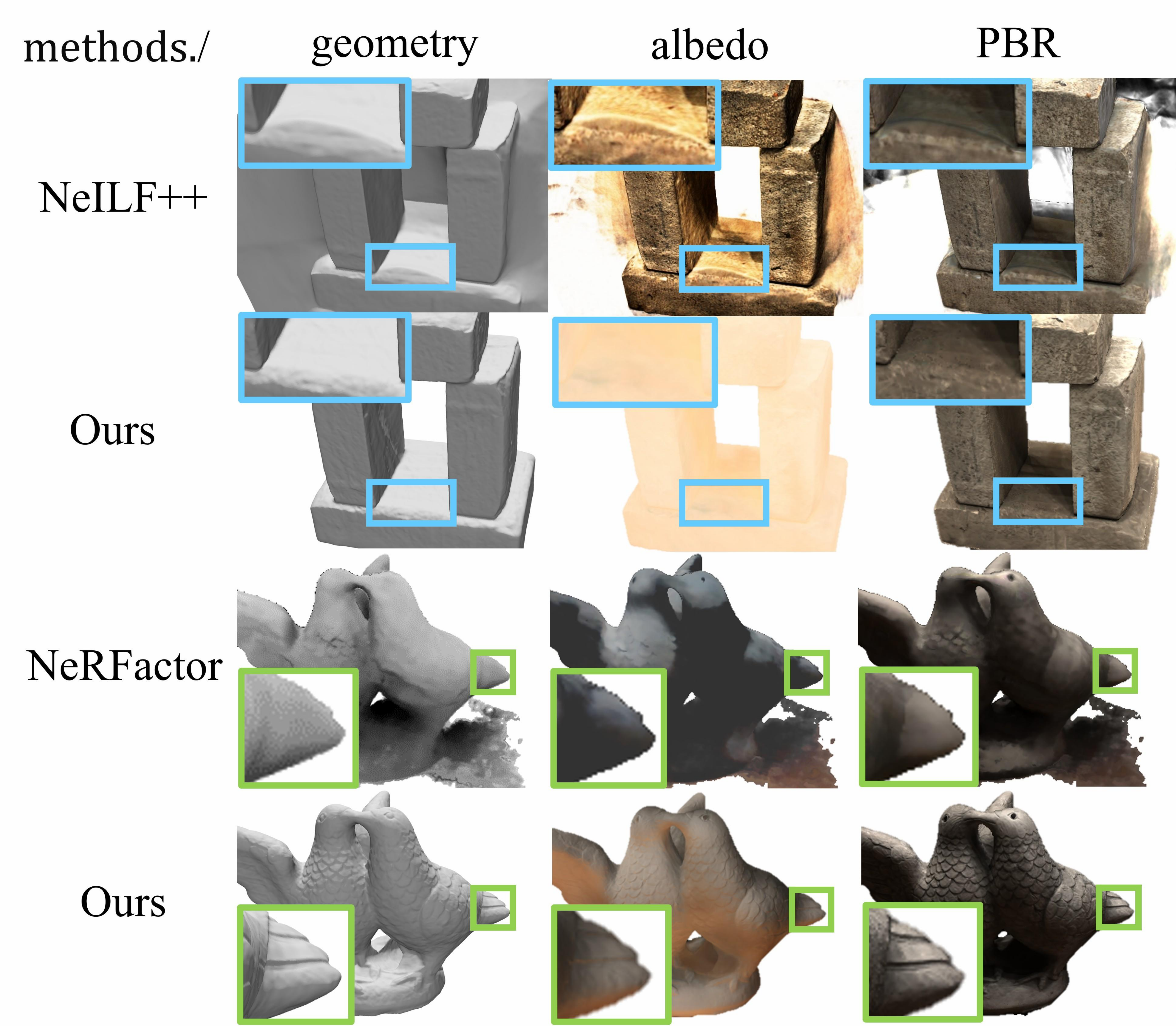}
\caption{Qualitative study on accumulated errors by comparing our method with the SOTA methods. Our method not only addresses the accumulated errors that occur in the previous methods, but also yields a more plausible albedo.}
\label{fig:errors}
\end{figure}

The results in the \cref{fig:wosurf} suggest that without the surface loss, the surface PBR, due to the nature of volume rendering, won't provide us with the desired results. While the \cref{tab:wovol} proves that supervising the volume rendering of inverse rendering field under the ground truth can help regularize the density field by considering the specular effects and textures.\\
\\
\textbf{Subsurface scattering: }We also test the role subsurface scattering plays in our framework. Although we do not have a ground truth of illumination for comparison, we choose DTU scan40~\cite{jensen2014large} for experiments because the framework predicts no specular effect on this scan so that the light directions are solely affected by the diffuse part. \cref{fig:sss} shows that with subsurface scattering, the framework can demonstrate more complex lighting environments. Intuitively, the white square and the yellow square in the images, though share the same normals, should have different lighting directions due to the occlusion and shades. \cref{fig:sss} proves that with subsurface scattering, the framework can tell such difference while for the traditional Lambertian model, the lighting directions are only dependent on the surface normals due to our MLP design.\\
\textbf{Accumulated errors: }To visualize the improvements of our method on alleviating the accumulated errors, we compared NePF with NeILF++~\cite{zhang2023neilf++} and NeRFactor~\cite{nerfactor} in ~\cref{fig:errors}.\\

\section{Limitations}
\label{sec:limitations}

While NePF ties the state-of-the-art methods in a single-stage manner, it relies on specific assumptions that limit complex lighting modeling. Future research might explore enhancing the robustness of our current coordinate-based lighting model. Additionally, there is still room for further advancements in geometry recovery, as our method finds difficulties in cases where the geometry variation is significant.
\section{Conclusions}
\label{sec:conclusions}

In this paper, we've proposed Neural Photon Field (NePF), a novel single-stage framework for the traditional multi-pass inverse rendering. By fully employing the physical meaning of the weight function of neural implicit surfaces and the outgoing radiance, we are able to recover all the desired properties in a single pass. Furthermore, we develop a novel coordinate-based illumination model that enhances the proposed one-pass technique. Our method is evaluated on both real-world and synthetic datasets. The results demonstrate that the proposed method obtains high-fidelity geometry, high-quality novel view synthesis, and alleviates the problem of accumulated errors and high computational waste in the traditional methods. We consider this work as a significant advancement in the field of inverse rendering.

{
    \small
    \bibliographystyle{ieeenat_fullname}
    \bibliography{main}
}

\clearpage
\setcounter{page}{1}
\maketitlesupplementary

\section {Network Structure}
\begin{figure}[h]
\centering
\includegraphics[width=1\linewidth]{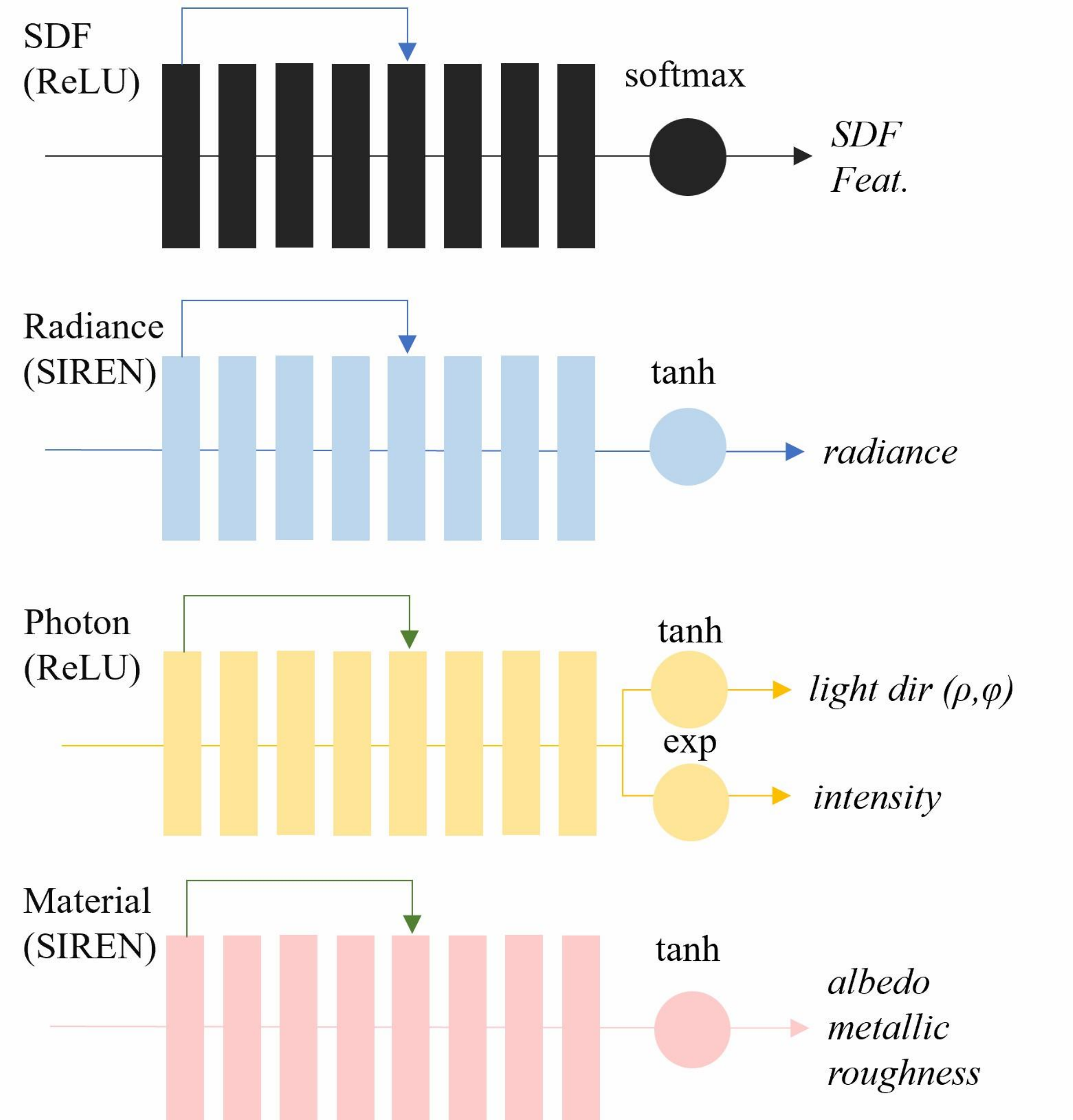}
\caption{Detailed network structure.}
\label{fig:detailnetwork}
\end{figure}

We use 8 layers SIREN~\cite{sitzmann2019siren} MLP with 256 width for outgoing radiance network and material network, while the photon network and the SDF network are relu activated with a depth of 8 and a width of 256. Detailed architecture is presented in ~\cref{fig:detailnetwork}

\section {Trainning scheme}
\subsection {Loss functions}
We use several loss functions during the trainning.
\subsubsection{Color loss}
The color loss has three parts that have been well described in ~\cref{sec:methods}. 
\begin{equation}
    \mathcal{L} = \mathcal{L}_r + \lambda _1 * \mathcal{L}_{surf}  + \lambda _2 * \mathcal{L}_{vol} .
\end{equation}
\subsubsection{Geo loss}
The geometry loss has two parts: The eikonal loss and the hessian loss.\\
\textbf{Eikonal loss} This is a typical loss function for SDF trainning that encourages the expectation of gradient to be 1.
\begin{align}
L_{eikonal} = | 1 - \| \nabla  f(x) \| | .
\end{align}
\textbf{Hessian loss} This loss function encourages a smoother reconstructed surface by minimizing the Hessian Matrix ~\cite{zhang2022critical}, which is the second derivative of the SDF. 
\begin{align}
L_{hessian} = | \nabla^2 f(x) | .
\end{align}
\subsubsection{Light variance loss}
For a smoother light direction prediction, we implement a light variance loss that impose a penalty on highly spatially-various illumination prediction.
\begin{align}
L_{light} = &\lambda_{3} * | var(x) - var(I) |+ \\
            &\lambda_{4} * | var(n) - var(l)| + \\
            &\lambda_{5} * |var(x) - var(l)| .
\end{align}
We use $x$ for coordinates, $n$ for surface normals, $I$ for light intensities and $l$ for light directions. The aim of this loss function is to assume that the illumination should encompass a local consistency. We assume that the variance of coordinates affect the variance of intensities, while both the variance of coordinates and the surface normals will affect the variance of light directions, though the former contribute more. We set $\lambda_{3}$ and $\lambda_{4}$ as 1.0 while $\lambda_{5}$ as 0.1. \\
The weight for the overall loss is tiny during the trainning. We use 0.0001 to prevent it affecting the overall trainning quality.
\subsubsection{Mask loss}
We run our method under the mask supervision. We use the typical BCE loss for mask supervision:
\begin{align}
 L_{mask} = BCE(weight_{sum}, mask) .
 \end{align}
\subsection{Learning Rate}
We implement the same anneal strategy of learning rate in NeuS~\cite{wang2021neus}. However, we set our base learning rate as 0.0003 for stability.
\subsection{Specular Terms}
We implement Schlick PBR~\cite{schlick} to model the specular part in BSDF estimation. For the Fresnel term:
\begin{align}
F_0 & = 0.04 \cdot (1 - m) + c \cdot m , \\
F(v,h) & = F_0 + (1 - F_0) (1 - h \cdot v)^5 .
\end{align}
For the geometry term:
\begin{align}
G1(v) & = \frac{n \cdot v}{2 \cdot \left[{\alpha + (1 - \alpha)(n \cdot v)}\right]} , \\
G2(l) & = \frac{n \cdot l}{2 \cdot \left[{\alpha + (1 - \alpha)(n \cdot l)}\right]} , \\
G(l, v) & = G1(v) \cdot G2(l).
\end{align}
And for the normal distribution term:
\begin{align}
D(h)= \frac{\alpha^2}{\pi \cdot ((n \cdot h)^2 (\alpha^2 - 1) + 1)^2}.
\end{align}
The $\alpha$ in the above equations are defined as the square of roughness.
\section{Synchronization}
\begin{figure}[t]
\centering
\includegraphics[width=1\linewidth]{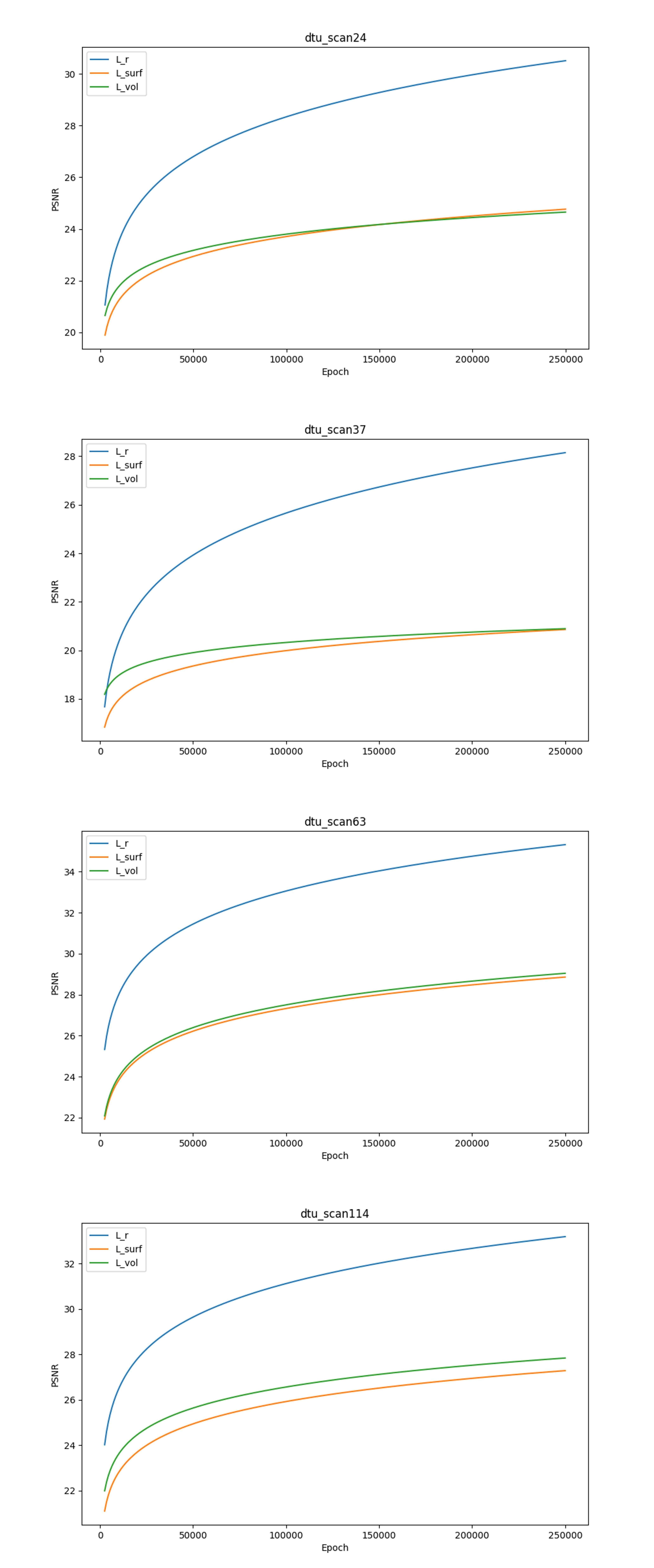}
\caption{Demonstrations of synchronization during the training process. The horizontal axis represents the epoch, while the vertical axis denotes the PSNR. The blue, orange, and green lines correspond to $L_r$, $L_{surf}$, $L_{vol}$ respectively. Four cases, scan24, scan37, scan63, and scan114 from DTU datasets~\cite{jensen2014large} are illustrated from top to bottom.}
\label{fig:sync3}
\end{figure}

To visualize the synchronization during the trainning process, we validate the PSNR of rendering results of outgoing radiance field, inverse rendering field and surface PBR every 2500 epoch in \cref{fig:sync3}

\section{Relighting and Material Editing}
\label{sec:mer}
\begin{figure*}[t]
\centering
\includegraphics[width=1\linewidth]{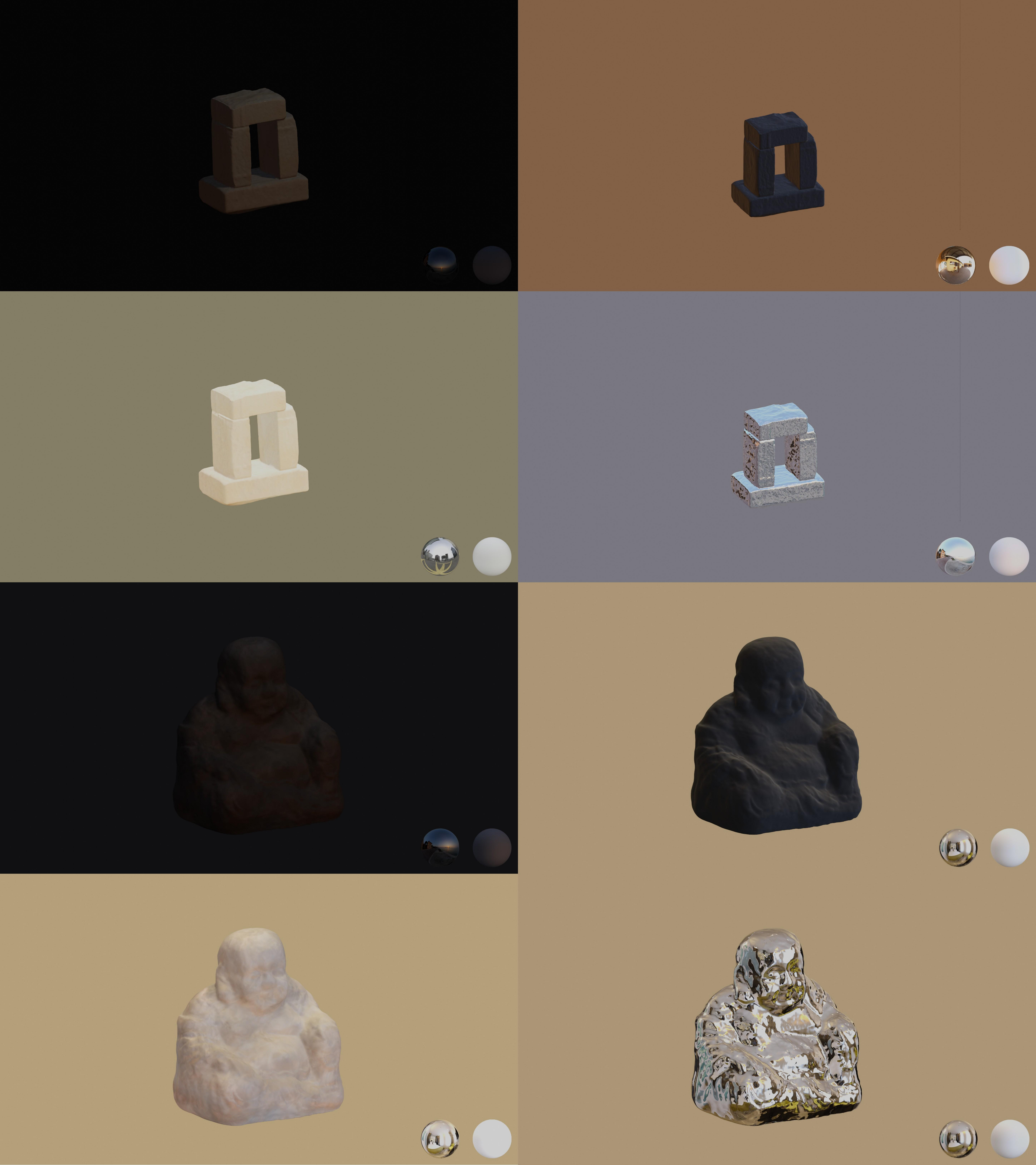}
\caption{Demonstrations of relighting and material editing effects. The left column is relighting, while the right column is material editing.}
\label{fig:materialedit}
\end{figure*}

We extract the meshes as well as its BSDF properties on the vertex and import them in Blender~\cite{blender} to demonstrate relighting and material editing ~\cref{fig:materialedit}.

%


\end{document}